\documentclass[11pt]{article}

\usepackage[preprint]{acl}

\usepackage{times}
\usepackage{latexsym}

\usepackage[T1]{fontenc}
\usepackage[utf8]{inputenc}

\usepackage{microtype}

\usepackage{inconsolata}

\usepackage{graphicx}

\usepackage{algorithm}
\usepackage{algorithmic}
\usepackage{times}
\usepackage{helvet}
\usepackage{courier}
\usepackage{booktabs}       % professional-quality tables

\usepackage{xcolor}         % colors
\usepackage{colortbl} 
\usepackage{arydshln} % 必须放在 colortbl 后面！

\usepackage{multirow}
\usepackage{amsmath,amssymb}
\usepackage{diagbox}
\usepackage{mathrsfs}
\usepackage{makecell}
\usepackage{enumitem}
\usepackage[english]{babel}
\usepackage{rotating}
\title{LACE-SVD: Loss-Aware SVD with Cumulative Error Correction for LLM Compression}

\author{%
\begin{tabular}{c@{\hspace{1.4em}}c@{\hspace{1.4em}}c@{\hspace{1.4em}}c@{\hspace{1.4em}}c@{\hspace{1.4em}}c}
\textbf{Zhuowen Liu}\textsuperscript{1} &
\textbf{Longkun Hao}\textsuperscript{2} &
\textbf{Shiyu Feng}\textsuperscript{3} &
\textbf{Xiaowen Chang}\textsuperscript{4} &
\textbf{Ruiqun Li}\textsuperscript{5} &
\textbf{Changqun Li}\textsuperscript{6\textdagger} \\
\end{tabular}\\[0.6em]
\begin{tabular}{c}
\textsuperscript{1}The Chinese University of Hong Kong \quad
\textsuperscript{2}Beihang University \\
\textsuperscript{3}Shanghai Jiao Tong University \\
\textsuperscript{4}University of Chinese Academy of Sciences \\
\textsuperscript{5}Beijing Institute of Technology \quad
\textsuperscript{6}East China Normal University
\end{tabular}%
}

\begin{document}
\maketitle
\begin{abstract}
The rapid growth in the parameter scale of large language models (LLMs) has
created a strong demand for efficient compression techniques. As a
hardware-agnostic and highly compatible approach, low-rank compression has been
widely adopted to reduce both memory footprint and computational cost. However, existing SVD-based methods are still largely driven by local
reconstruction objectives, overlooking two critical limitations: rank budgets
are often allocated without explicitly considering layer-wise loss sensitivity,
and local approximation errors can propagate and accumulate through the residual
stream, leading to amplified global deviations from the original model. To address
these issues, we propose \textbf{LACE-SVD}, a \textbf{L}oss-\textbf{A}ware SVD
framework with \textbf{C}umulative \textbf{E}rror correction for LLM
compression. LACE-SVD first estimates the calibration negative-log-likelihood
increase induced by candidate layer-wise compression ratios and solves a
budget-constrained allocation problem to assign rank budgets. It then refines
the compressed model with closed-form local updates and introduces a
propagation-aware correction for residual-stream output modules, reducing
layer-output discrepancy as a proxy for cumulative error propagation. Experimental results demonstrate that at a high compression ratio (0.6), the WikiText-2 PPL of our method on LLaMA-7B (32.57) is significantly better than that of Dobi-SVD (46.18).
\end{abstract}

\section{Introduction}

Large language models (LLMs), including GPT Family~\cite{achiam2023gpt,dettmers2022gpt3}, LLaMA Family~\cite{touvron2023llama,dubey2024llama}, and Qwen Family~\cite{bai2023qwen,yang2025qwen3} have achieved remarkable performance across commonsense reasoning, question answering, document understanding, and few-shot learning. However, their strong capabilities come with substantial memory and computational costs, making billion-parameter models difficult to deploy under practical latency, memory, and energy constraints. To address this challenge, post-training compression has become an important research direction, including quantization~\cite{frantar2022gptq,huostquant}, pruning~\cite{slicegpt}, knowledge distillation~\cite{hinton2015distilling}, and low-rank decomposition~\cite{golub1987generalization,asvd-yuan2023asvd}.
Among these approaches, singular value decomposition (SVD) based compression is
particularly attractive because it directly factorizes dense weight matrices
into low-rank components, reducing both model size and matrix multiplication
cost without requiring large-scale retraining.

Recent SVD-based compression methods~\cite{asvd-yuan2023asvd,li2025adasvd,svd-llm-wangsvd} improve over vanilla truncated SVD by
using activation statistics or calibration data. These methods rescale or
whiten weight matrices before decomposition so that the low-rank approximation
better matches the activation distribution of each layer. Although effective,
they are still mainly driven by local reconstruction objectives. That is, they
aim to preserve the output of each compressed module or layer under calibration
activations, while the global language modeling behavior is only indirectly
considered.

This local view leads to two limitations. First, different layers exhibit
different sensitivity to compression. A uniform rank ratio may over-compress
loss-critical layers and under-compress robust layers, resulting in a
suboptimal use of the global parameter budget. Second, reconstruction errors are
not isolated across layers. Once an early layer is compressed, its output
perturbation enters the residual stream and changes the input distribution of
subsequent layers. These perturbations can accumulate through attention and MLP
blocks, causing the compressed model to deviate from the original model
even when each individual layer has a small local reconstruction error.

Motivated by these observations, we propose \textbf{LACE-SVD}, a
\textbf{L}oss-\textbf{A}ware SVD framework with \textbf{C}umulative
\textbf{E}rror correction for LLM compression. LACE-SVD addresses the above two
problems from complementary perspectives. First, it performs loss-aware
layer-wise rank allocation by evaluating how candidate layer-wise compression
ratios affect calibration negative log-likelihood, and then selecting one ratio
per layer under a global parameter budget. This allows the method to allocate
more capacity to layers that are more important for language modeling quality. Second, LACE-SVD introduces cumulative error correction after SVD
decomposition. We refine the low-rank factors with closed-form local updates
using full-precision calibration activations. Furthermore, for modules that
directly write to the residual stream, such as attention output projections and
MLP down projections, we introduce a propagation-aware correction mechanism.
This correction does not directly optimize the language modeling loss; instead,
it reduces the discrepancy between the compressed layer output and the
original layer output, serving as a practical proxy for mitigating
cumulative error propagation. An acceptance gate keeps the correction only when
it improves layer-level output fidelity on calibration data. We empirically demonstrate that our method significantly outperforms Dobi-SVD at high compression ratios, achieving 32.57 WikiText-2 PPL on LLaMA-7B (0.6 ratio) compared to Dobi-SVD's 46.18. In summary, LACE-SVD shifts post-training SVD compression from purely local
matrix approximation toward loss-aware budget allocation and cumulative error
correction. Our contributions are as follows:

\begin{itemize}
    \item We identify that existing SVD-based LLM compression methods are
    limited by heuristic rank allocation and cumulative error propagation across
    layers, showing that minimizing local reconstruction error alone is
    insufficient for preserving end-to-end model fidelity.

    \item We propose a loss-aware layer-wise rank allocation strategy that uses
    calibration negative log-likelihood to select compression ratios under a
    global parameter budget, allocating more capacity to layers that are more
    critical to language modeling performance.

    \item We introduce cumulative error correction through propagation-aware
    local updates for residual-stream output modules, reducing layer-output
    discrepancy between the compressed and original models without
    requiring full end-to-end fine-tuning.
\end{itemize}

\section{Related Works}

\textbf{Large Language Model Compression.} LLM compression broadly falls into pruning, quantization, knowledge distillation, and low-rank approximation. Unstructured and structured pruning \citep{slicegpt,zhang2023loraprune,ma2023llm, dettmers2024spqr} often struggle with a trade-off between hardware efficiency and severe accuracy degradation. Quantization \citep{nagel2021white,frantar2022gptq,huostquant,lin2024awq,hu2024I-llm} delivers strong memory savings but loses flexibility and performance under aggressive low-bit settings, while knowledge distillation \citep{hinton2015distilling} requires costly retraining. In contrast, low-rank approximation via Singular Value Decomposition (SVD) \citep{asvd-yuan2023asvd, svd-llm-wangsvd} offers a post-training, hardware-agnostic solution that is naturally orthogonal to other compression paradigms.

\textbf{SVD-based Compression for LLMs.} SVD \citep{golub1987generalization} reduces matrix sizes by truncating the smallest singular values. Low-rank approximation using SVD has been widely studied as an efficient approach for compressing LLMs. Early works mainly focused on minimizing per-layer truncation loss. To incorporate parameter importance, FWSVD \citep{fwsvd-hsulanguage} introduces Fisher information to weigh the importance of parameters, but its complex gradient calculation demands substantial computing and memory resources. To address the impact of activation distributions, ASVD \citep{asvd-yuan2023asvd} scales the weight matrix with a diagonal matrix to normalize the impact of input channels. SVD-LLM \citep{svd-llm-wangsvd} makes further advancements by proposing a data-whitening strategy on the input matrix to mitigate its impact, achieving a theoretical minimal truncation loss. Building upon this, SVD-LLM v2~\cite{svd-llm-v2-wang-etal-2025-svd-llm} further optimizes the truncation process by dynamically assigning unique, layer-wise compression ratios based on theoretical loss. Furthermore, Dobi-SVD~\cite{dobi-svd-wang2025dobi} tackles the challenges of layer-wise rank allocation and information loss by introducing a differentiable truncation mechanism for adaptive rank search. To address broader architectural issues, SAES-SVD~\cite{saes-svd-hu2026saes} tackles the problem of cross-layer error accumulation by introducing a cumulative error-aware compression objective alongside an adaptive weighting mechanism to dynamically suppress both local and propagated errors.

\begin{figure*}[htb] % htbp 表示排版位置首选here(当前位置)，其次top(页首)等
    \centering
    % 如果您的pdf放在了名为 figures/ 的文件夹下，请写 figures/compression_ratio_analysis.pdf
    \includegraphics[width=\linewidth]{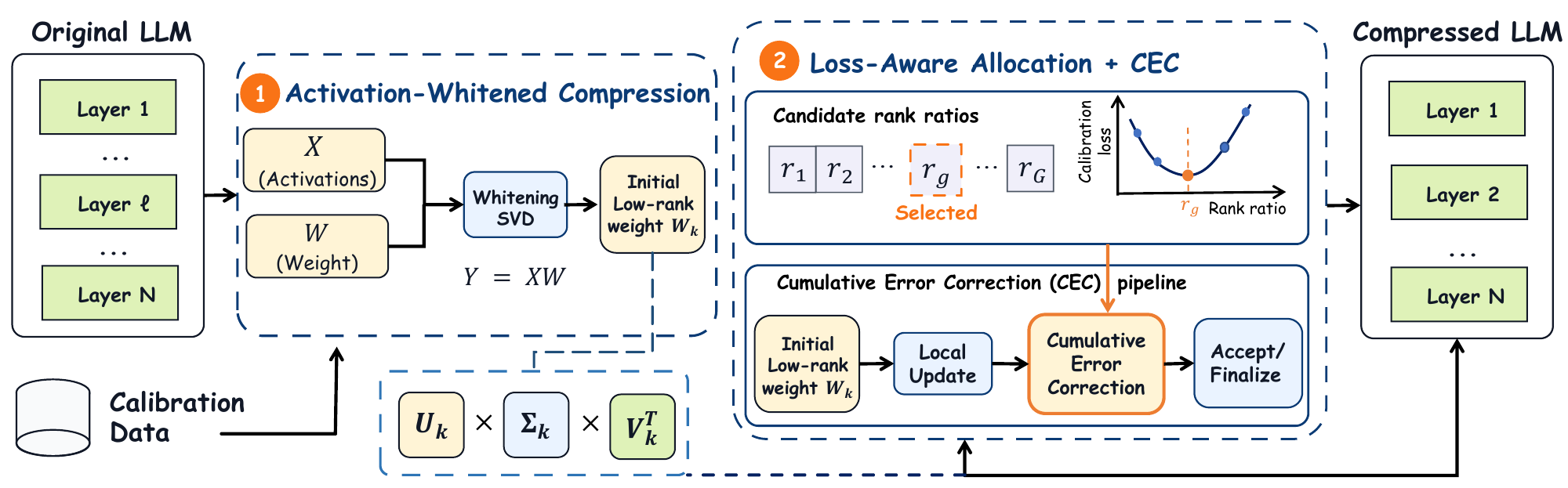}
    \caption{\textbf{Overall architecture of LACE-SVD.} Original LLM layers are first decomposed into initial low-rank weights using Activation-Whitened SVD. A loss-aware allocation strategy then dynamically selects the optimal rank ratio ($r_g$) for each layer. Finally, an error correction pipeline, incorporating Local Updates and Cumulative Error Correction (CEC), is applied to compensate for the cumulative errors before assembling the Compressed LLM.}
    \label{Overall}
\end{figure*}

\begin{figure*}[htb] % htbp 表示排版位置首选here(当前位置)，其次top(页首)等
    \centering
    % 如果您的pdf放在了名为 figures/ 的文件夹下，请写 figures/compression_ratio_analysis.pdf
    \includegraphics[width=\linewidth]{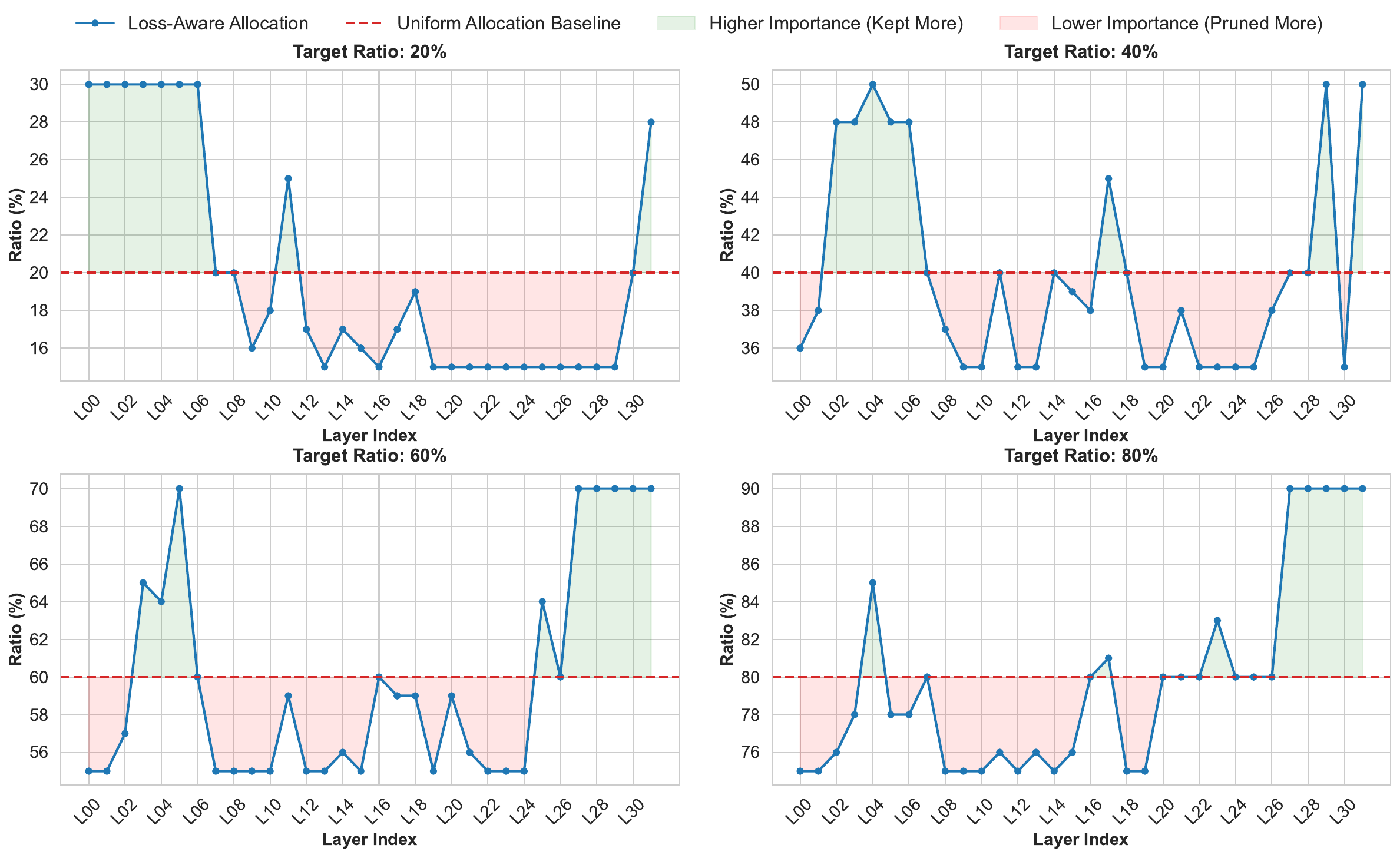}
    \caption{\textbf{Layer-wise compression ratio allocation for LLaMA-7B under different targets.} The loss-aware allocation strategy dynamically assigns compression budgets across 32 layers. Compared to the uniform baseline (red dashed line), the algorithm consistently preserves more parameters in the early layers (L00–L06, green region) to maintain foundational lexical and syntactic features, as well as in the final layers (L27–L31), which act as a highly sensitive bottleneck crucial for vocabulary alignment. Conversely, the intermediate layers (L08–L26, red region) are aggressively pruned due to greater semantic and rank redundancy. This distinct U-shaped “preserve-ends, prune-middle” pattern remains robustly consistent across all target compression constraints (20\% to 80\%).}
    \label{fig:compression_allocation}
\end{figure*}

\section{Method}

\subsection{Overview}

Given a pretrained LLM $f_{\theta}$ with Transformer layers
$\{F_\ell\}_{\ell=1}^{L}$, our goal is to obtain a compressed model
$f_{\hat{\theta}}$ under a target global compression ratio $\rho$, while preserving the
behavior of the full-precision model. LACE-SVD follows a post-training
compression setting and does not require end-to-end fine-tuning.

The method consists of three main stages. First, we compute activation-whitened
low-rank decompositions for linear projection matrices using calibration
activations. Second, we perform loss-aware layer-wise rank allocation by
estimating the calibration loss increase caused by different candidate
compression ratios for each layer. Third, after decomposing the selected layers,
we refine the low-rank factors with closed-form local updates and further apply
cumulative error correction to residual-stream output modules. The algorithm description is provided in Appendix~\ref{algorithm_lace_svd}.

\subsection{Activation-Whitened SVD}

For a linear projection with weight matrix
$W \in \mathbb{R}^{d_{\mathrm{out}} \times d_{\mathrm{in}}}$ and calibration
inputs $X \in \mathbb{R}^{n \times d_{\mathrm{in}}}$, vanilla truncated SVD
minimizes the reconstruction error of $W$ itself. However, in an LLM, the effect
of approximating $W$ depends on the distribution of its input activations. We
therefore use an activation-whitened decomposition.

We first estimate the input covariance:
\begin{equation}
    G = X^\top X.
\end{equation}
Let $C$ be the Cholesky factor of $G$, i.e.,
\begin{equation}
    G \approx CC^\top .
\end{equation}
We then decompose the activation-scaled weight matrix:
\begin{equation}
    WC = U \Sigma V^\top .
\end{equation}
Keeping the top-$k$ singular components gives the compressed approximation:
\begin{equation}
    \widehat{W}
    =
    U_k \Sigma_k V_k^\top C^{-1}.
\end{equation}

This corresponds to the profiling matrix path in the implementation: calibration
activations are accumulated to build the whitening matrix, and the SVD is
performed on the scaled weight matrix before mapping the approximation back to
the original weight space.

\subsection{Loss-Aware Layer-Wise Rank Allocation}

Uniformly assigning the same compression ratio to all layers ignores the fact that
layers have different sensitivity to compression. LACE-SVD therefore selects
layer-wise compression ratios using calibration negative log-likelihood.

Let $\mathcal{R}_{\rho}$ denote the candidate keep-ratio set associated with the
target budget $\rho$. For each layer $\ell$ and candidate ratio
$r \in \mathcal{R}_{\rho}$, we temporarily compress only layer $\ell$ while
keeping the other layers unchanged. We then evaluate the calibration loss
increase:
\begin{equation}
    \Delta_{\ell,r}
    =
    \mathcal{L}_{\mathrm{calib}}
    \left(f_{\theta}^{(\ell,r)}\right)
    -
    \mathcal{L}_{\mathrm{calib}}
    \left(f_{\theta}\right),
\end{equation}
where $f_{\theta}$ is the full-precision model and
$f_{\theta}^{(\ell,r)}$ denotes the model whose $\ell$-th layer is compressed
with ratio $r$.

The implementation evaluates candidate ratios by temporarily replacing the
weights in a layer with their low-rank reconstructed dense weights, measuring the
calibration loss, and then restoring the original weights. The resulting
per-layer table contains the tuple
\begin{equation}
    (r, C_{\ell,r}, \Delta_{\ell,r}),
\end{equation}
where $C_{\ell,r}$ is the parameter cost of compressing layer $\ell$ with ratio
$r$.

Given these tables, LACE-SVD solves a multiple-choice budget allocation problem:
\begin{equation}
    \min_{\{r_\ell\}_{\ell=1}^{L}}
    \sum_{\ell=1}^{L} \Delta_{\ell,r_\ell},
\end{equation}
subject to
\begin{equation}
    \sum_{\ell=1}^{L} C_{\ell,r_\ell}
    \leq
    \rho C_{\mathrm{full}},
    \quad
    r_\ell \in \mathcal{R}_{\rho}.
\end{equation}
Here $C_{\mathrm{full}}$ is the total number of parameters before compression.
This objective explicitly connects rank allocation with language-modeling
degradation, allowing the method to allocate more rank budget to loss-sensitive
layers.

In the implementation, this optimization is solved by a discretized dynamic
programming procedure. For efficiency, candidate evaluation can be performed in
two stages: a coarse screening stage selects promising ratios for each layer,
and a full calibration evaluation stage computes the final loss table. This is an
engineering acceleration and does not change the allocation objective.

\subsection{Simultaneous Local Closed-Form Update}

After selecting layer-wise compression ratios, we replace each target projection with a
low-rank factorization. For a compressed projection, write
\begin{equation}
    \widehat{W} = U V,
\end{equation}
where $V$ is the input-side factor and $U$ is the output-side factor. Given
calibration input $X$ and the corresponding full-precision output
\begin{equation}
    Y = X W^\top,
\end{equation}
we define the low-rank hidden feature
\begin{equation}
    Z = X V^\top.
\end{equation}
The output-side factor is then updated by solving a regularized least-squares
problem:
\begin{equation}
    U^{\star}
    =
    \arg\min_U
    \left\|
    ZU^\top - Y
    \right\|_F^2
    +
    \lambda_U
    \left\|
    U - U_0
    \right\|_F^2,
\end{equation}
where $U_0$ is the initial factor obtained from SVD and $\lambda_U$ is a ridge
regularization coefficient.

The implementation accumulates the normal equations over calibration batches and
solves this update in closed form. It also supports a weighted variant:
\begin{equation}
\resizebox{\linewidth}{!}{$\displaystyle
    U^{\star}
    =
    \arg\min_U
    \left\|
    D^{1/2}
    \left(
    ZU^\top - Y
    \right)
    \right\|_F^2
    +
    \lambda_U
    \left\|
    U - U_0
    \right\|_F^2,
$}
\end{equation}

where $D$ is a diagonal sample-weight matrix computed from the selected weighting
mode. In the main configuration, the residual-based weighting mode is used.

LACE-SVD uses a simultaneous local update mode. Specifically, the calibration
statistics for modules in a layer are collected using full-precision layer
activations, and the low-rank factors are solved after the statistics are
accumulated. This avoids immediately feeding partially compressed intermediate
states into later modules during calibration, which can otherwise make the local
targets unstable.

The implementation also optionally evaluates a bi-side closed-form correction,
where both the output-side and input-side factors are considered. However, the
candidate correction is accepted only if it improves the held-out reconstruction
objective by at least a minimum relative gain. Thus, the method does not assume
that the bi-side update is always beneficial; it selects the safer update
according to calibration reconstruction error.

\subsection{Cumulative Error Correction}

While local reconstruction objectives effectively reduce module-level errors, they do not fully address cumulative error propagation. In Transformer architectures, specific projection modules write directly back to the residual stream. Errors within these modules inevitably alter the hidden states consumed by all subsequent layers, leading to compounded degradation across the network. 

To mitigate this, we define a subset of target modules, denoted as $\mathcal{S}$, within each Transformer layer. LACE-SVD introduces a propagation-aware cumulative error correction mechanism applied to the modules in $\mathcal{S}$. For a given layer $\ell$ and a target module $s \in \mathcal{S}$, let $M_{\ell,s}^{\mathrm{full}}$ represent the output of the full-precision module, and $M_{\ell,s}^{\mathrm{cmp}}$ denote the output of the compressed module under the same calibration input. We construct a propagation-aware target as follows:
\begin{equation}
\label{eq:pa_target}
    T_{\ell,s}^{\mathrm{pa}}
    =
    M_{\ell,s}^{\mathrm{cmp}}
    +
    \alpha
    \left(
    M_{\ell,s}^{\mathrm{full}}
    -
    M_{\ell,s}^{\mathrm{cmp}}
    \right),
\end{equation}
where $\alpha \in [0,1]$ is a hyperparameter that controls the correction strength. An ablation study on $\alpha$ is provided in Appendix~\ref{sec:ablation_alpha}.

This target moves the compressed residual-stream contribution toward the
full-precision contribution. Importantly, this step does not directly optimize
the language modeling loss. Instead, it uses module and layer output discrepancy
as a practical proxy for cumulative error propagation. After constructing
$T_{\ell,s}^{\mathrm{pa}}$, we re-solve the output-side factor of the selected
module using the same closed-form local update objective, replacing $Y$ with
$T_{\ell,s}^{\mathrm{pa}}$.

To prevent over-correction, LACE-SVD applies a layer-level acceptance gate. Let
$H_\ell$ be the calibration input to layer $\ell$,
$F_\ell^{\mathrm{full}}$ be the full-precision layer,
$F_\ell^{\mathrm{cmp}}$ be the compressed layer before propagation-aware
correction, and $F_\ell^{\mathrm{pa}}$ be the corrected layer. The correction is
accepted only if
\begin{equation}
\resizebox{\linewidth}{!}{$
    \left\| F_\ell^{\mathrm{pa}}(H_\ell) - F_\ell^{\mathrm{full}}(H_\ell) \right\|_F^2
    <
    \left\| F_\ell^{\mathrm{cmp}}(H_\ell) - F_\ell^{\mathrm{full}}(H_\ell) \right\|_F^2 .
$}
\end{equation}

If this criterion is not satisfied, the propagation-aware update is rejected and
the previous compressed factors are restored. This gate ensures that cumulative
error correction is retained only when it improves layer-level fidelity.

\section{Experiments}
\textbf{Models and Datasets.}\quad To comprehensively evaluate the performance of LACE-SVD across different models and tasks, we conducted systematic experiments on mainstream open-source LLMs using standard language understanding and generation benchmarks, covering a wide range of parameter budgets. By default, all results are obtained by applying a single-pass SVD layer by layer, without any extra task-specific fine-tuning after compression. For the model selection, we chose well-known architectures, including three different LLM families at  different scales (LLaMA-7B, 13B, LLaMA2-7B~\cite{touvron2023llama2}, OPT-6.7B~\cite{zhang2022opt}, Vicuna-7B~\cite{chiang2023vicuna} and Mistral-7B~\cite{jiang2023mistral}. Regarding the evaluation, we mainly focused on natural language reasoning and understanding. Specifically, to assess the language coherence and modeling capacity of the compressed models, we calculated the perplexity on the WikiText2~\cite{merity2016pointer} and C4~\cite{raffel2020exploring} datasets with a sequence length of 2048. We also measured zero-shot accuracy on several datasets—including ARC-Challenge~\cite{clark}, ARC-Easy~\cite{clark}, HellaSwag~\cite{zellers-etal-2019-hellaswag}, MathQA~\cite{amini-etal-2019-mathqa}, PIQA~\cite{bisk2020piqa}, and WinoGrande~\cite{sakaguchi2019winogrande}—to examine how well the models generalize across a variety of tasks.

\textbf{Baselines.}\quad Under a unified calibration and evaluation protocol, we compared our method against several well-established SVD-based low-rank compression approaches, such as ASVD~\cite{asvd-yuan2023asvd}, SVD-LLM~\cite{svd-llm-wangsvd}, FWSVD~\cite{fwsvd-hsulanguage}, Dobi-SVD~\cite{dobi-svd-wang2025dobi}. In addition to SVD methods, we compared our work with other types of LLM compression approaches. Specifically, these include three state-of-the-art pruning-based methods: LLM-Pruner~\cite{ma2023llm}, SliceGPT~\cite{slicegpt}, and BlockPruner~\cite{zhong2025blockpruner}. Implementation Details is provided in Appendix~\ref{implementation_details}.

% Additionally, based on empirical observations, we restricted the target module subset for cumulative error correction to $\mathcal{S} = \{\texttt{o\_proj}, \texttt{down\_proj}\}$.

\textbf{Main Results.} We conducted a comprehensive evaluation of our proposed method (LACE-SVD) for low-rank compression of LLMs and compared it with several strong SVD-based baselines across different compression ratios. Table 1 summarizes the perplexity and zero-shot evaluation results on the representative LLaMA-7B. Overall, LACE-SVD consistently achieves superior performance. At the highly challenging 0.6 compression ratio, LACE-SVD exhibits remarkable robustness. While existing advanced methods suffer severe degradation—with SVD-LLM and Dobi-SVD yielding WikiText-2 perplexities of 53.74 and 46.18, respectively—our method maintains a significantly lower perplexity of 32.57. Beyond perplexity, LACE-SVD effectively minimizes the degradation in downstream capabilities across varying ratios. For instance, at the 0.4 compression ratio, LACE-SVD restricts the average zero-shot accuracy drop to 23.1\%, which is substantially lower than those of Dobi-SVD (26.9\%) and SVD-LLM (28.9\%).

\begin{table*}[htbp]
\centering
\caption{Perplexity and zero-shot evaluation of LLaMA-7B across seven benchmark datasets under varying compression ratios. The table compares LACE-SVD with competing SVD-based methods (ASVD$^*$, FWSVD, SVD-LLM , Dobi-SVD). Methods with fine-tuning are marked by $^\dagger$.}
\label{main_experiments}
  \resizebox{\textwidth}{!}{
    \begin{tabular}{llrr|rrrrrrrrr}
    \toprule
    \textbf{Ratio} & \textbf{Method} & \textbf{Wiki2}$\downarrow$ &  \textbf{C4}$\downarrow$ & \textbf{Openb.}$\uparrow$ & \textbf{ARC\_e}$\uparrow$ & \textbf{ARC\_c}$\uparrow$ & \textbf{WinoG.}$\uparrow$ & \textbf{HellaS.}$\uparrow$ & \textbf{PIQA}$\uparrow$ & \textbf{MathQA}$\uparrow$ & \textbf{Avg.}$\uparrow$ & \textbf{Drop}$\downarrow$ \\
    \midrule
    \textbf{0.0} & Baseline & 5.68   & 7.34  & 0.28  & 0.67  & 0.38  & 0.67  & 0.56  & 0.78  & 0.27  & 0.52  & 0.00 \\
    \midrule
    \multirow{6}[2]{*}{\textbf{0.2}} & FWSVD & 2e5     & 2e3   & 0.09  & 0.11  & 0.06  & 0.05  & 0.08  & 0.10  & 0.05  & 0.08  & 84.6\% \\
          & ASVD$^\dagger$ & 11.14 & 15.93 & 0.25  & 0.53  & 0.27  & 0.64  & 0.41  & 0.68  & 0.24  & 0.43  & 17.3\% \\
          & SVD-LLM & 7.94  & 15.84 & 0.22  & 0.58  & 0.29  & 0.63  & 0.43  & 0.69  & 0.24  & 0.44  & 15.4\% \\
          & Dobi-SVD & 8.54  & {10.01} & 0.26  & 0.59  & 0.31  & {0.66} & {0.44}  & 0.70  & 0.23  & 0.46  & 11.5\% \\
          & \textbf{LACE-SVD} & \textbf{7.39} & \textbf{10.99} & \textbf{0.26} & \textbf{0.62} & \textbf{0.33}  & \textbf{0.66} & \textbf{0.45} & \textbf{0.71} & \textbf{0.24} & \textbf{0.47} & \textbf{9.6\%} \\
    \midrule
    \multirow{6}[2]{*}{\textbf{0.4}} & FWSVD & 2e4     & 1e4   & 0.06  & 0.05  & 0.02  & 0.02  & 0.00  & 0.05  & 0.03  & 0.03  & 94.2\% \\
          & ASVD$^\dagger$ & 1e3     & 1e3   & 0.13  & 0.28  & 0.22  & 0.48  & 0.26  & 0.55  & 0.19  & 0.30  & 42.3\% \\
          & SVD-LLM & 13.11 & 49.83 & 0.19  & 0.42  & 0.25  & 0.58  & 0.33  & 0.60  & 0.21  & 0.37  & 28.9\% \\
          & Dobi-SVD & 13.54 & 23.54 & {0.22}  & 0.41  & \textbf{0.27}  & 0.58  & {0.34}  & 0.61  & {0.23}  & 0.38  & 26.9\% \\
          & \textbf{LACE-SVD} & \textbf{12.00} & \textbf{21.30} & \textbf{0.23} & \textbf{0.50} & 0.26  & \textbf{0.60} & \textbf{0.35} & \textbf{0.62} & \textbf{0.23} & \textbf{0.40} & \textbf{23.1\%} \\
    \midrule
    \multirow{5}[2]{*}{\textbf{0.6}} & FWSVD & 3e4     & 2e4   & 0.06  & 0.01  & 0.00  & 0.00  & 0.01  & 0.01  & 0.00  & 0.01  & 98.1\% \\
          & ASVD$^\dagger$ & 6e4     & 4e5   & 0.12  & 0.26  & 0.21  & 0.49  & 0.26  & 0.53  & 0.18  & 0.29  & 44.2\% \\
          & SVD-LLM & 53.74   & 345.49   & 0.14  & 0.28  & 0.22  & 0.50  & 0.27  & 0.55  & 0.21  & 0.31  & 40.4\% \\
          & Dobi-SVD & 46.18   & 190.62   & {0.15}  & 0.31  & {0.20}  & 0.52  & 0.28  & 0.54  & 0.22  & {0.32}  & {38.4\%} \\
          & \textbf{LACE-SVD} & \textbf{32.57} & \textbf{72.15} & \textbf{0.16} & \textbf{0.32} & \textbf{0.23} & \textbf{0.52} & \textbf{0.29} & \textbf{0.55} & \textbf{0.22} & \textbf{0.33} & \textbf{36.5\%} \\
    \midrule
    \multirow{2}[2]{*}{\textbf{0.8}} & SVD-LLM & 1349     & 6224   & 0.07  & 0.03  & 0.01  & 0.04  & 0.02  & 0.07  & 0.01  & 0.04  & 92.3\% \\
          & \textbf{LACE-SVD} & \textbf{238.05} & \textbf{406.50} & \textbf{0.14} & \textbf{0.27} & \textbf{0.20} & \textbf{0.51} & \textbf{0.26} & \textbf{0.53} & \textbf{0.21} & \textbf{0.30} & \textbf{42.3\%} \\
    \bottomrule
    \end{tabular}}
\end{table*}

\textbf{Performance on Different LLM Architectures.} To evaluate the generalizability of our proposed LACE-SVD across diverse model architectures, we benchmark it against baseline methods on four distinct models from different LLM families: OPT-6.7B, LLaMA-2-7B, Mistral-7B, and Vicuna-7B. All models are evaluated under a 20\% compression ratio on WikiText-2 (perplexity) and six common sense reasoning datasets (average accuracy). As demonstrated in Table~\ref{different_llms}, vanilla SVD and FWSVD suffer from catastrophic collapse across all four models, yielding practically unusable perplexities and near-zero zero-shot accuracies. Notably, compared to the strongest baseline SVD-LLM, our LACE-SVD achieves lower perplexity on every model. For instance, on Mistral-7B and Vicuna-7B, LACE-SVD reduces the perplexity to 8.46 and 7.91, respectively, significantly outperforming SVD-LLM. Furthermore, LACE-SVD consistently achieves the highest average accuracy among all compressed models, further validating its superior ability to preserve the downstream reasoning capabilities of various LLMs.

\begin{table*}[t]
\centering
\caption{Perplexity ($\downarrow$) of LACE-SVD and baselines on WikiText-2 and the average accuracy ($\uparrow$) of the six common sense reasoning datasets of four different LLMs-OPT-6.7B, LLaMA-2-7B, Mistral-7B, and Vicuna-7B--under 20\% compression ratio.}
\label{different_llms}
\resizebox{\textwidth}{!}{
\begin{tabular}{c|cc|cc|cc|cc}
\toprule
 & \multicolumn{2}{c}{{OPT-6.7B}} & \multicolumn{2}{c}{{LLaMA 2-7B}} & \multicolumn{2}{c}{{Mistral-7B}} & \multicolumn{2}{c}{{Vicuna-7B}} \\ \midrule
{Method} & \textbf{Perplexity}$\downarrow$ & \textbf{Accuracy}$\uparrow$ & \textbf{Perplexity}$\downarrow$ & \textbf{Accuracy}$\uparrow$ & \textbf{Perplexity}$\downarrow$ & \textbf{Accuracy}$\uparrow$ & \textbf{Perplexity}$\downarrow$ & \textbf{Accuracy}$\uparrow$ \\ \midrule
\textcolor{gray}{Original} & \textcolor{gray}{10.86} & \textcolor{gray}{0.52} & \textcolor{gray}{5.47} & \textcolor{gray}{0.57} & \textcolor{gray}{5.25} & \textcolor{gray}{0.61} & \textcolor{gray}{6.78} & \textcolor{gray}{0.56} \\ \midrule
SVD & 66275 & 0.03 & 18192 & 0.09 & 159627 & 0.03 & 18644 & 0.05 \\
FWSVD & 14559 & 0.06 & 2360 & 0.12 & 6357 & 0.08 & 2758 & 0.09 \\
ASVD & 82.00 & 0.32 & 10.10 & 0.36 & 13.72 & 0.32 & 16.23 & 0.33 \\ 
SVD-LLM & {16.04}  & {0.41}  & {8.50}  & {0.53}  & {10.21}  & {0.42}  & {8.41}  & {0.51}  \\ \midrule
LACE-SVD & \textbf{15.39}  & \textbf{0.43}  & \textbf{7.54}  & \textbf{0.55}  & \textbf{8.46}  & \textbf{0.50}  & \textbf{7.91}  & \textbf{0.54}  \\ \bottomrule
\end{tabular}
}
\end{table*}

\begin{figure}[htb]
    \centering
    % 左边的独立图 (Figure X)
    \begin{minipage}[b]{0.45\textwidth}
        \centering
        \includegraphics[width=\textwidth]{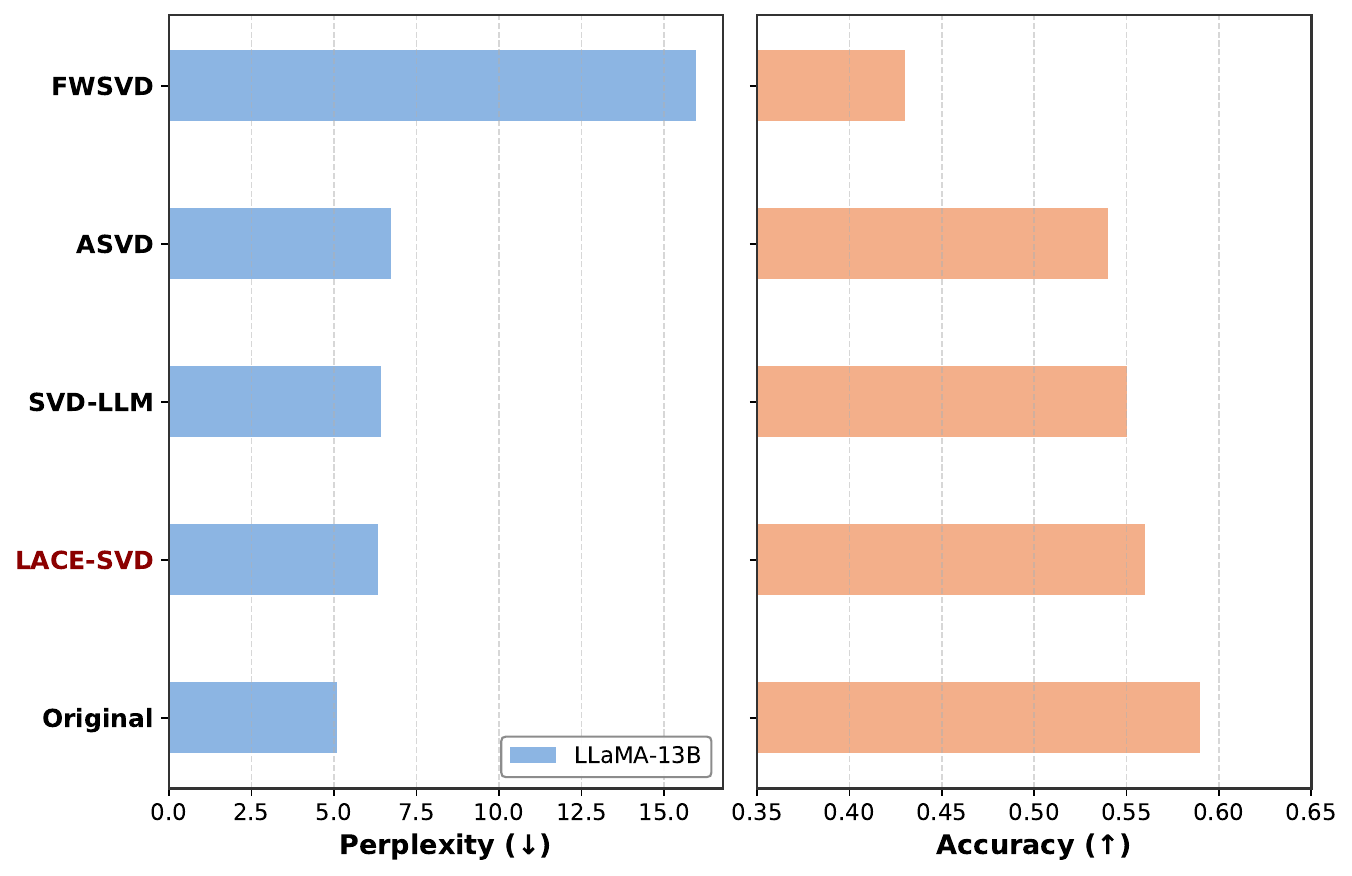}
        \caption{Perplexity ($\downarrow$) and average accuracy ($\uparrow$) of LLaMA-13B under 20\% compression.}
        \label{flarge_llm}
    \end{minipage}
    \hfill % 添加弹性间距
    % 右边的独立图 (Figure X+1)
    \begin{minipage}[b]{0.45\textwidth}
        \centering
        \includegraphics[width=\textwidth]{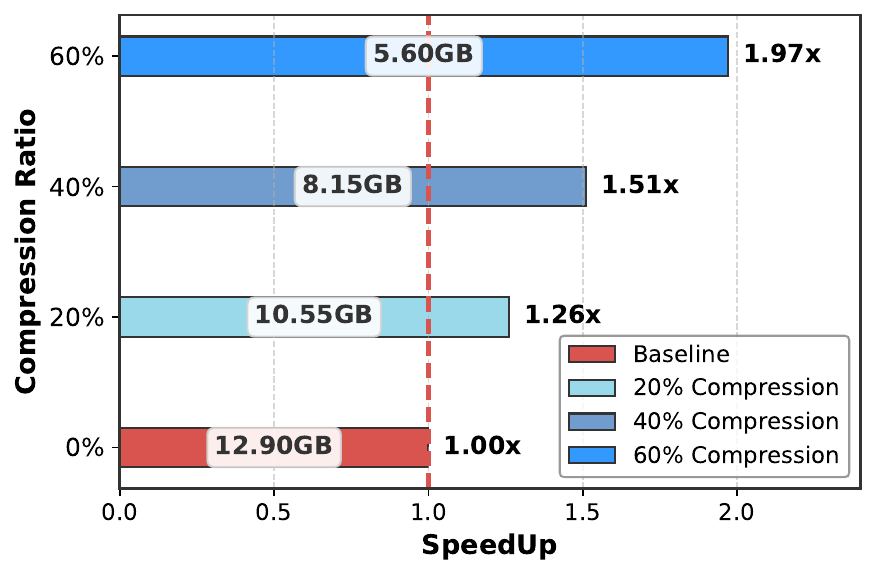}
        \caption{Memory usage and inference speedup of LLaMA-7B on varying compression ratios.}
        \label{speedup}
    \end{minipage}
    
\end{figure}

\textbf{Performance on LLMs with Larger Scales.} To examine the generalizability and scalability of our proposed method, we evaluate its performance on larger language models, specifically LLaMA-13B, under a 20\% compression ratio. As shown in Fig~\ref{flarge_llm}, LACE-SVD consistently outperforms all baseline methods.

\textbf{Memory Usage and Inference Speedup.} Low-rank approximation inherently decreases both computational complexity and parameter storage, leading to a dual reduction in memory footprint and inference latency. We evaluated the inference speedup and memory consumption of our proposed LACE-SVD on LLaMA-7B under varying compression ratios on a single NVIDIA H200 GPU. As illustrated in Figure~\ref{speedup}, LACE-SVD achieves consistent and significant inference acceleration over the uncompressed baseline. At a 20\% compression ratio, the model achieves a 1.26$\times$ speedup while reducing the memory requirement from the baseline's 12.90 GB to 10.55 GB. As the compression ratio increases, the efficiency gains become even more pronounced. At a 40\% compression ratio, the speedup reaches 1.51$\times$ with an 8.15 GB memory footprint. Most notably, at an aggressive 60\% compression ratio, LACE-SVD delivers nearly a factor of two acceleration (1.97$\times$ speedup) while slashing the memory demand by more than half, down to just 5.60 GB. These strictly aligned results strongly demonstrate the practicality of LACE-SVD in facilitating highly efficient LLM deployment in resource-constrained environments.

\textbf{Comparison with Structured Pruning.} We also evaluate our proposed LACE-SVD against three state-of-the-art structured pruning-based LLM compression methods: LLM-Pruner~\cite{ma2023llm}, SliceGPT~\cite{slicegpt}, and BlockPruner~\cite{zhong2025blockpruner}. The comparison is conducted on LLaMA-7B using the WikiText-2 dataset under strictly identical memory budgets, ranging from 10 GB down to 7 GB. As shown in Table~\ref{pruning_wikitext}, LACE-SVD consistently and significantly outperforms all three structured pruning baselines across all memory constraints. More importantly, while the pruning-based methods suffer from severe performance degradation as the memory budget becomes tighter, our method exhibits remarkable robustness. For instance, when the memory budget drops to 8 GB and 7 GB, the perplexities of the pruning methods soar to the range of 16.39 to 43.05. In stark contrast, LACE-SVD maintains remarkably low and stable perplexities of 11.27 and 18.81, respectively. In particular, under the highly restrictive 7 GB memory budget, LACE-SVD achieves a 13\% reduction in perplexity even when compared to the best-performing pruning baseline, LLM-Pruner (18.81 vs. 21.68), demonstrating the superior memory efficiency and preservation ability of our low-rank approach.

\begin{table}[t]
\centering
\caption{WikiText-2 perplexity of LLaMA-7B against structured pruning across different memory budgets.}
\label{pruning_wikitext}

\small
\setlength{\tabcolsep}{7pt}
\begin{tabular}{c|c|c|c|c}
\toprule
\multicolumn{1}{c}{} & \multicolumn{4}{c}{Under Various Memory Budgets} \\
\midrule
Method & 10 GB & 9 GB & 8 GB & 7 GB \\
\midrule
LLM-Pruner & 9.88 & 12.21 & 18.94 & 21.68 \\
SliceGPT & 8.78 & 12.73 & 16.39 & 27.41 \\
BlockPruner & 9.40 & 12.76 & 19.78 & 43.05 \\
\midrule
\textbf{LACE-SVD} & \textbf{7.60} & \textbf{8.85} & \textbf{11.27} & \textbf{18.81} \\
\bottomrule
\end{tabular}

\end{table}

\subsection{Ablation Study on Core Components}

We conduct a systematic ablation study under a fixed $0.6$ compression ratio to isolate the contributions of LACE-SVD's core components over the foundational Activation-Whitened SVD baseline, denoted as SVD-LLM (Table \ref{tab:ablation_components}). At this aggressive compression level, the baseline suffers severe performance degradation, making these structural enhancements strictly necessary. \textbf{Loss-Aware Rank Allocation:} Replacing uniform rank assignment with loss-aware allocation improves WikiText-2 PPL from 53.74 to 45.18. By explicitly tying allocation to language-modeling degradation, it efficiently directs the parameter budget to loss-sensitive layers, preventing catastrophic information bottlenecking in critical transformer blocks. \textbf{Simultaneous Local Update:} Applying the closed-form local update to the baseline yields a clear gain, reducing PPL from 53.74 to 43.56. Solving the regularized least-squares problem minimizes module-level errors, while our simultaneous mode avoids calibration instabilities caused by partially compressed intermediate states. This ensures the reconstructed activations faithfully match the full-precision outputs before progressing deeper into the network. \textbf{Cumulative Error Correction (CEC):} Adding CEC over the local update further reduces PPL to 38.25. This confirms that adjusting residual-stream projections, guarded by a layer-level acceptance gate, successfully mitigates the compounded degradation that local objectives alone cannot resolve. \textbf{Full LACE-SVD Synergy:} Combining all components achieves the best performance (PPL: 32.57). The allocation module provides an optimal parameter foundation, while the combined correction pipeline systematically compensates for truncation gaps at both module and network levels. Ultimately, our unified framework reduces the PPL by over 21 points compared to the baseline, demonstrating that structural allocation and multi-granularity error correction are highly synergistic.

\begin{table}[h]
\centering
\caption{\textbf{Ablation study on the core components of LACE-SVD.} We report the Perplexity (PPL, lower is better) on WikiText-2 under 0.6 compression ratio. \textit{SVD-LLM} denotes the foundational Activation-Whitened SVD.}
\label{tab:ablation_components}
\resizebox{\columnwidth}{!}{
\begin{tabular}{l | c}
\toprule
\textbf{Method Variant} & \textbf{Wikitext-2 PPL} $\downarrow$  \\
\midrule
SVD-LLM (Baseline)  & 53.74 \\
\midrule
SVD-LLM + Allocation  & 45.18 \\
SVD-LLM + Local Update  & 43.56 \\
SVD-LLM + Local Update + CEC &  38.25 \\
\midrule
\textbf{Full LACE-SVD} & \textbf{32.57} \\
\bottomrule
\end{tabular}
}
\end{table}

\section{Conclusion}
In this paper, we introduced LACE-SVD to address the suboptimal parameter distribution and error accumulation issues in existing SVD-based LLM compression. It bridges the gap between local matrix factorization and global model fidelity through two mechanisms: loss-aware layer-wise rank allocation and propagation-aware cumulative error correction. Experiments show our approach maintains strong language modeling performance and significantly outperforms strong baselines under high compression ratios, proving its effectiveness for resource-constrained LLM deployment.

\section*{Limitations}
While LACE-SVD is highly effective for post-training LLM compression, it involves two practical trade-offs: First, our rank allocation evaluates layer sensitivities independently. Although this ignores non-linear cross-layer coupling, this necessary approximation ensures computational tractability and enables an efficient dynamic programming solution. Second, the cumulative error correction uses layer-output discrepancy ($L_2$ distance) as a proxy instead of directly optimizing the global loss. This design deliberately avoids the prohibitive costs of end-to-end backpropagation, striking a practical balance between compression efficiency and model fidelity.

% \section*{Acknowledgments}

\bibliography{custom}

\newpage
\appendix

\section{Algorithm}
\label{algorithm_lace_svd}
\begin{algorithm}[htbp]
\caption{LACE-SVD: Loss-Aware and Cumulative Error Corrected SVD}
\label{alg:lace_svd}
\begin{algorithmic}[1]
\renewcommand{\algorithmicrequire}{\textbf{Input:}}
\renewcommand{\algorithmicensure}{\textbf{Output:}}
\REQUIRE Pretrained LLM $f_{\theta}$ with layers $\{F_\ell\}_{1}^{L}$, calibration data $X$, target budget $\rho$, candidate ratios $\mathcal{R}_{\rho}$, correction strength $\alpha$, regularization $\lambda_U$.
\ENSURE Compressed LLM $f_{\hat{\theta}}$

\vspace{0.05in}
% \STATE \emph{\% Initialization}
\STATE Compute Activation-Whitened SVD for all weights to obtain initial factors.

\vspace{0.05in}
\STATE \textbf{Contribution 1: Loss-Aware Layer-Wise Rank Allocation}
\FOR{each layer $\ell \in \{1, \dots, L\}$ and candidate ratio $r \in \mathcal{R}_{\rho}$}
    \STATE Evaluate loss gap: $\Delta_{\ell,r} \gets \mathcal{L}_{\mathrm{calib}}\big(f_{\theta}^{(\ell,r)}\big) - \mathcal{L}_{\mathrm{calib}}(f_{\theta})$
\ENDFOR
\STATE Solve DP: $\{r_\ell^\star\}_{1}^{L} \gets \arg\min \sum \Delta_{\ell,r_\ell} \quad \text{s.t.} \quad \sum C_{\ell,r_\ell} \leq \rho C_{\mathrm{full}}$

\vspace{0.05in}
\STATE \textbf{Contribution 2: Local Update \& Cumulative Error Correction}
\FOR{each layer $\ell = 1 \dots L$}
    \FOR{each target projection $W$ in $F_\ell$}
        \STATE Truncate factors to rank $r_\ell^\star$ to get $U_0$ and $V$.
        \STATE Hidden feature $Z \gets XV^\top$, full-precision target $Y \gets XW^\top$.
        
        \STATE \emph{Simultaneous Local Closed-Form Update}
        \STATE $U^{\star} \gets \arg\min_U \| ZU^\top - Y \|_F^2 + \lambda_U \| U - U_0 \|_F^2$
        
        \STATE \emph{Cumulative Error Correction (CEC)}
        \IF{module $s \in \mathcal{S} = \{\texttt{o\_proj}, \texttt{down\_proj}\}$}
            \STATE Target $T_{\ell,s}^{\mathrm{pa}} \gets M_{\ell,s}^{\mathrm{cmp}} + \alpha (M_{\ell,s}^{\mathrm{full}} - M_{\ell,s}^{\mathrm{cmp}})$
            \STATE Re-solve for $U_{\mathrm{pa}}^{\star}$ replacing $Y$ with $T_{\ell,s}^{\mathrm{pa}}$.
            \STATE Let $F_\ell^{\mathrm{pa}}$ and $F_\ell^{\mathrm{cmp}}$ be layers using $U_{\mathrm{pa}}^{\star}$ and $U^{\star}$ respectively.
            \IF{$\| F_\ell^{\mathrm{pa}}(H_\ell) - F_\ell^{\mathrm{full}}(H_\ell) \|_F^2 < \| F_\ell^{\mathrm{cmp}}(H_\ell) - F_\ell^{\mathrm{full}}(H_\ell) \|_F^2$}
                \STATE Accept correction: $U^{\star} \gets U_{\mathrm{pa}}^{\star}$
            \ENDIF
        \ENDIF
        
        \STATE Update compressed projection: $\widehat{W} \gets U^{\star}V$
    \ENDFOR
\ENDFOR

\RETURN $f_{\hat{\theta}}$ defined by updated $\{\widehat{W}\}$
\end{algorithmic}
\end{algorithm}

\section{Implementation Details and Main Hyperparameters.}
\label{implementation_details}
To ensure a fair comparison, we followed SVD-LLM and randomly selected 256 samples from WikiText-2 as calibration data.  All experiments were executed on multiple NVIDIA H200 GPUs. 

\begin{table}[h]
\centering
\small
\begin{tabular}{ll}
\toprule
Item & Value \\
\midrule
Backbone & LLaMA-7B \\
Target compression ratio & $\rho=0.4$ \\
Calibration corpus & Wikitext-2 \\
Whitening samples & 256 \\
Whitening sequence length & 2048 \\
Loss-aware batches & 64 \\
Loss-aware sequence length & 1024 \\
Loss-aware batch size & 16 \\
Candidate evaluation & two-stage \\
Stage-1 batches & 16 \\
Stage-1 top-$k$ & 4 \\
DP budget bins & 4000 \\
Local update mode & simultaneous \\
Local update samples & 64 \\
Local update micro-batch size & 8 \\
Minimum held-out gain & $2 \times 10^{-4}$ \\
Propagation-aware modules & \texttt{o\_proj}, \texttt{down\_proj} \\
Propagation strength & $\alpha=0.7$ \\
Propagation gate batches & 64 \\
\bottomrule
\end{tabular}
\caption{Main hyperparameters for the LLaMA-7B $\rho=0.4$ setting.}
\end{table}

\section{Ablation Study on Correction Strength ($\alpha$)}
\label{sec:ablation_alpha}

\begin{figure}[htb]
    \centering
    % 导入由 Python 生成的 PDF 图片
    \includegraphics[width=0.7\linewidth]{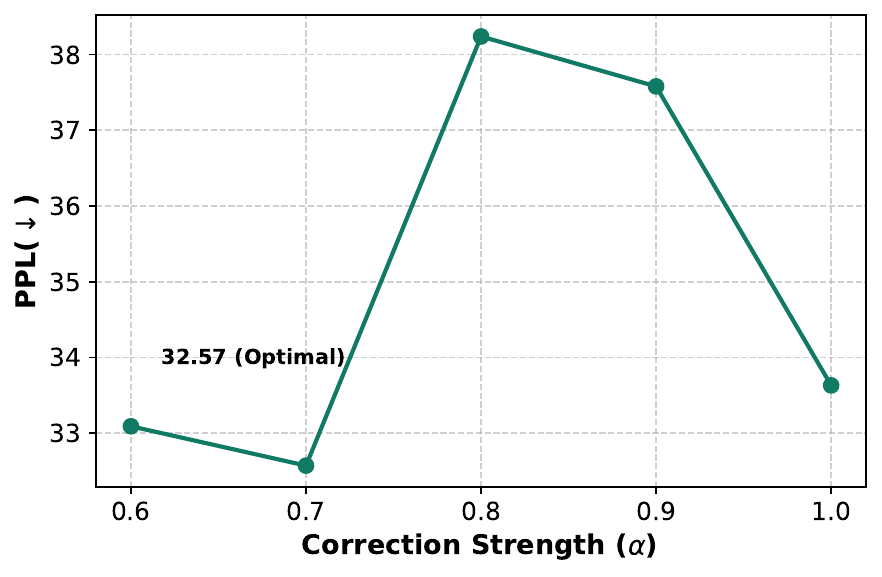}
    % 图片下方的 Caption
    \caption{Ablation study of the correction strength $\alpha$ on WikiText-2 perplexity using the LLaMA-7B model. The compression ratio (0.6) is kept identical across all settings.}
    \label{alpha_ablation_llama7b}
\end{figure}

In the LACE-SVD local update stage, we employ a propagation-aware cumulative error correction mechanism for modules in $\mathcal{S}$. As formulated in Equation~\ref{eq:pa_target}, the target activation $T_{\ell,s}^{\mathrm{pa}}$ is constructed by interpolating between the full-precision module output $M_{\ell,s}^{\mathrm{full}}$ and the compressed module output $M_{\ell,s}^{\mathrm{cmp}}$ using the correction strength $\alpha \in [0, 1]$. To investigate the impact of $\alpha$ on the final language modeling performance, we conduct an ablation study on the WikiText-2 dataset across various values. The perplexity (PPL) trends are illustrated in Figure~\ref{alpha_ablation_llama7b}.

As observed in Figure~\ref{alpha_ablation_llama7b}, the performance exhibits a highly non-linear relationship with $\alpha$. Specifically, the framework achieves the best perplexity (\textbf{32.57}) when $\alpha=\textbf{0.7}$. In this regime, the target blends the full-precision output with a moderate proportion of the current $M_{\ell,s}^{\mathrm{cmp}}$, acting as a strong implicit regularizer that restricts the magnitude of the matrix update. This prevents aggressive overfitting and preserves the baseline orthogonality of the truncated SVD. Interestingly, a counter-intuitive performance spike (severe degradation) occurs when $\alpha$ is slightly increased to the range of $[\textbf{0.8}, \textbf{0.9}]$, peaking at over \textbf{38} for $\alpha=\textbf{0.8}$. We attribute this degradation to the structural noise introduced into the closed-form solver: blending this specific, unbalanced ratio of the inherently biased compressed output into the target forces the solver to fit a distorted objective, leading to catastrophic error accumulation in deeper layers. Conversely, as $\alpha$ reaches $\textbf{1.0}$, the objective degenerates to $T_{\ell,s}^{\mathrm{pa}} = M_{\ell,s}^{\mathrm{full}}$, functioning as pure teacher forcing. While supplying the exact, uncompressed target activations allows the performance to recover significantly (yielding a PPL of approximately \textbf{33.6}), it remains sub-optimal compared to the interpolated target at $\alpha=\textbf{0.7}$. Based on these empirical findings, we fix $\alpha = \textbf{0.7}$ for all our main experiments to ensure optimal and stable layer-wise reconstruction.

\section{Calibration Setup}

We use Wikitext-2 as the calibration corpus. Whitening statistics are
computed from 256 calibration samples with sequence length 2048. For
loss-aware layerwise allocation, candidate ratios are evaluated using
64 calibration batches with sequence length 1024 and batch size 16.

\subsection{Candidate Ratio Search}

For each target compression ratio $\rho$, we construct a finite candidate set
$\mathcal{R}_\rho$ around $\rho$. The exact candidate set is treated as
an experimental hyperparameter and varies with the target compression
ratio. The final layerwise ratios are selected by solving the
multiple-choice knapsack problem.

\subsection{Two-Stage Candidate Evaluation}

Evaluating all candidates for all layers on the full calibration set is
expensive. We therefore use a two-stage candidate evaluation strategy.
In the first stage, all candidates are evaluated using 16 calibration
batches. For each layer, we retain the top candidates according to
calibration loss increase and cost-effectiveness. In the second stage,
the retained candidates are evaluated using the full calibration set.

This procedure reduces computational cost but does not change the final
budgeted optimization objective.

\subsection{Local Update Details}

After layerwise allocation, we perform simultaneous local update. All
linear modules in a layer are fitted using the original full-precision
activations. We use residual-based token weighting and a held-out
selection rule. The minimum relative held-out improvement is
$2 \times 10^{-4}$.

The ridge coefficients are:
\begin{equation}
    \lambda_U = 10^{-5},
    \qquad
    \lambda_V = 10^{-4}.
\end{equation}
The singular value floor used for numerical stability is $10^{-5}$.

\subsection{Propagation-Aware Refinement Details}

Propagation-aware residual local update is applied to
\texttt{o\_proj} and \texttt{down\_proj}. We use correction strength
$\alpha=0.7$ and evaluate the layer-output acceptance gate on at most
64 calibration batches per layer. Candidate updates that fail the gate
are discarded and the previous factors are restored.

\subsection{Efficiency Engineering}

We cache loss-aware candidate tables to avoid repeated candidate
evaluation. The cache stores layerwise candidate losses, parameter
costs, and metadata. We also optionally preload calibration batches onto
the GPU to reduce host-device transfer overhead. These components are
used only for efficiency and do not alter the compression objective.

\end{document}